\documentclass[conference]{IEEEtran}
\def\IEEEtitletopspace{0.63cm}
\IEEEoverridecommandlockouts
\usepackage[a4paper, left=1.91cm, right=1.16cm, top=3.67cm, bottom=1.91cm]{geometry}
\usepackage[font=small]{caption}
\usepackage{cite}
\usepackage{amsmath,amssymb,amsfonts}
\usepackage{algorithmic}
\usepackage{graphicx}
\usepackage{textcomp}
\usepackage{xcolor}
\usepackage{hyperref}
\usepackage{enumitem}
\usepackage{makecell}
\usepackage{array}
\usepackage[bottom,flushmargin]{footmisc}
\usepackage{subcaption}
\usepackage{gensymb}

\def\BibTeX{{\rm B\kern-.05em{\sc i\kern-.025em b}\kern-.08em
    T\kern-.1667em\lower.7ex\hbox{E}\kern-.125emX}}

\begin{document}

\title{MiNI-Q: A Miniature, Wire-Free Quadruped with Unbounded, Independently Actuated Leg Joints\\}

\author{%
Daniel Koh\textsuperscript{1,*}, %
Suraj Shah\textsuperscript{1,*}, %
Yufeng Wu\textsuperscript{1}, %
and Dennis Hong\textsuperscript{1}%
}
\maketitle

\begingroup
\renewcommand\thefootnote{1}
\footnotetext[1]{All authors are with the Department of Mechanical and Aerospace Engineering,
UCLA, Los Angeles, CA 90095, USA. \texttt{\{danielkoh1, surajnshah, ericyufengwu, dennishong\}@g.ucla.edu}.
\(^{*}\)Denotes equal contribution.}
\endgroup

\begin{abstract}
Physical joint limits are common in legged robots and can restrict workspace, constrain gait design, and increase the risk of hardware damage. This paper introduces MiNI-Q\footnotemark[2], a miniature, wire-free quadruped robot with independently actuated, mechanically unbounded 2-DOF leg joints. We present the mechanical design, kinematic analysis, and experimental validation of the proposed robot. The leg mechanism enables both oscillatory gaits and rotary locomotion while allowing the robot to fold to a minimum height of 2.5 cm. Experimentally, MiNI-Q achieves speeds up to 0.46 m/s and demonstrates low-clearance crawling, stair climbing, inverted locomotion, jumping, and backflipping. The wire-free architecture extends our previous Q8bot design \cite{Q8bot}, improving assembly reliability at miniature scale. All mechanical and electrical design files are released open source to support reproducibility and further research.
\end{abstract}

\begingroup
\renewcommand\thefootnote{2}
\footnotetext[2]{Project page: https://github.com/RoMeLaUCLA/miniq-public}
\endgroup

\section{Introduction}

Legged robots are suitable for navigating complex environments effectively due to their adaptability to unknown terrains \cite{QuadrupedReview}. Over the past decade, we have seen significant advancements in legged robotics, including novel leg mechanisms \cite{LegDisney, LegOncilla}, integrated actuators \cite{ActuatorBEAR}, and locomotion control \cite{RLReview}. 

The mechanisms used in multi-legged robots can be categorized into serial, parallel, and hybrid topologies \cite{MechanismReview}. Serial mechanisms mimic animal and human anatomy and have the advantages of simplified kinematics, compact size, and larger workspace. In contrast, parallel mechanisms are implemented for their increased stiffness and lower leg inertia, with the trade-off of reduced workspace and complex kinematics \cite{MrongaParallel}. Regardless of the topology, most existing legged robots have finite joint ranges that restrict mobility, as cables routed throughout the leg are subjected to strain and eventual failure due to large and repetitive joint movements. While parallel mechanisms mitigate this issue by keeping joint actuators stationary relative to the robot base \cite{xu2023design, Doggo}, mechanical hard stops remain common across both serial and parallel leg designs, which can damage the hardware when improper motion commands exceed the robot’s physical workspace.

Joint limits in some serial linkages can be extended through careful cable routing \cite{Lee_AiDIN,Grimminger_Solo}, but are difficult to eliminate entirely unless special components are used, such as slip rings \cite{BillingsBDPatent}, which adds significant cost and complexity to legged mechanisms. Parallel mechanisms, like a five-bar linkage, can eliminate joint limits through coaxial joint placement, a practice that is common in wheel--leg transformable robots that combine the advantages of both locomotion modalities \cite{Lywal,RHex-T3}. In these designs, circular rims are attached to the five-bar linkage, which forms a wheel-like profile at specific joint configurations. The coaxial placement of the actuators enables continuous joint rotation, allowing the robot to travel in wheeled mode on flat terrains. However, the underlying mechanism is biased toward wheeled operation, resulting in limited legged capability in reachable workspace and achievable gait variations. Stanford Doggo uses a similar coaxial drive assembly that allows unlimited rotation of its leg linkages \cite{Doggo}, but the associated advantages were not explicitly studied.

Other research has demonstrated the benefits of continuously rotating joints in purely legged robots. For example, work on NaBi explored a continuously rotating knee joint on a bipedal robot, which enabled the robot to step over doorsills, climb stairs, and traverse obstacles \cite{NaBi}. These results suggest that the benefits of continuous joint rotation extend beyond simply preventing mechanical damage; enhanced locomotion independent of any wheeled transformation is now viable for legged robots.

Building on these insights, we explore a legged robot mechanism in which both the shoulder and knee joints are independently actuated and unbounded. We realize our design through a miniature-scale robot, MiNI-Q, to validate the concept while keeping the potential to generalize to larger-sized applications. The main contributions of this paper are:

\begin{itemize}
    \item A compact, wire-free quadruped robot architecture as an evolution of the Q8bot design \cite{Q8bot}.
    \item A two-degree-of-freedom, belt-driven leg mechanism without physical joint limits.
\end{itemize}

\begin{figure}[t]
    \centering
    \includegraphics[width=0.48\textwidth]{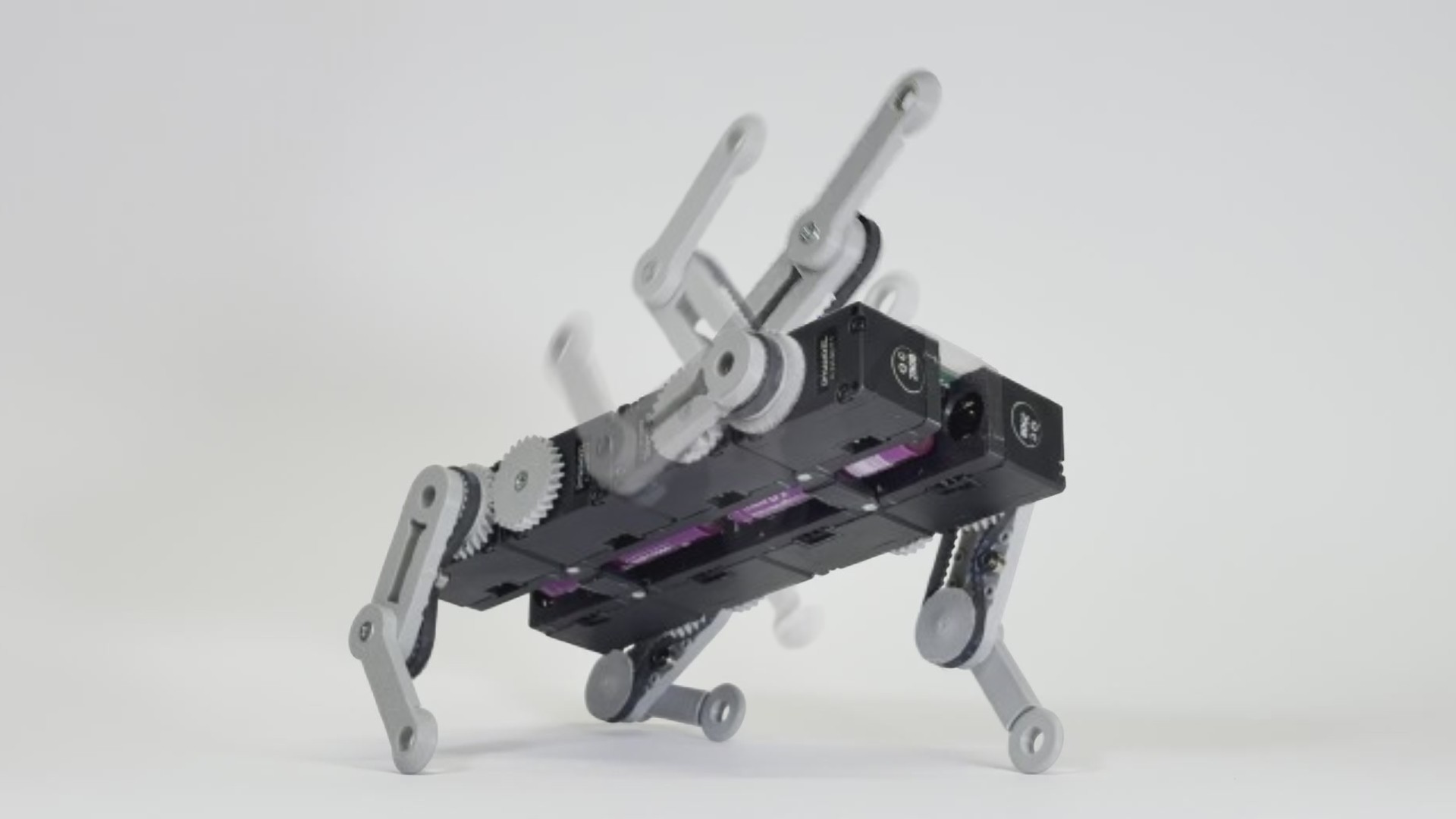}  
    \captionsetup{belowskip=0pt}
    \captionsetup{font=footnotesize}
    \caption{MiNI-Q, a miniature quadruped robot featuring a 2R leg mechanism designed to remove conventional joint limits.}
    \label{fig:heroshot}
\end{figure}

\begin{figure*}[t]  
    \centering
    \includegraphics[width=\textwidth]{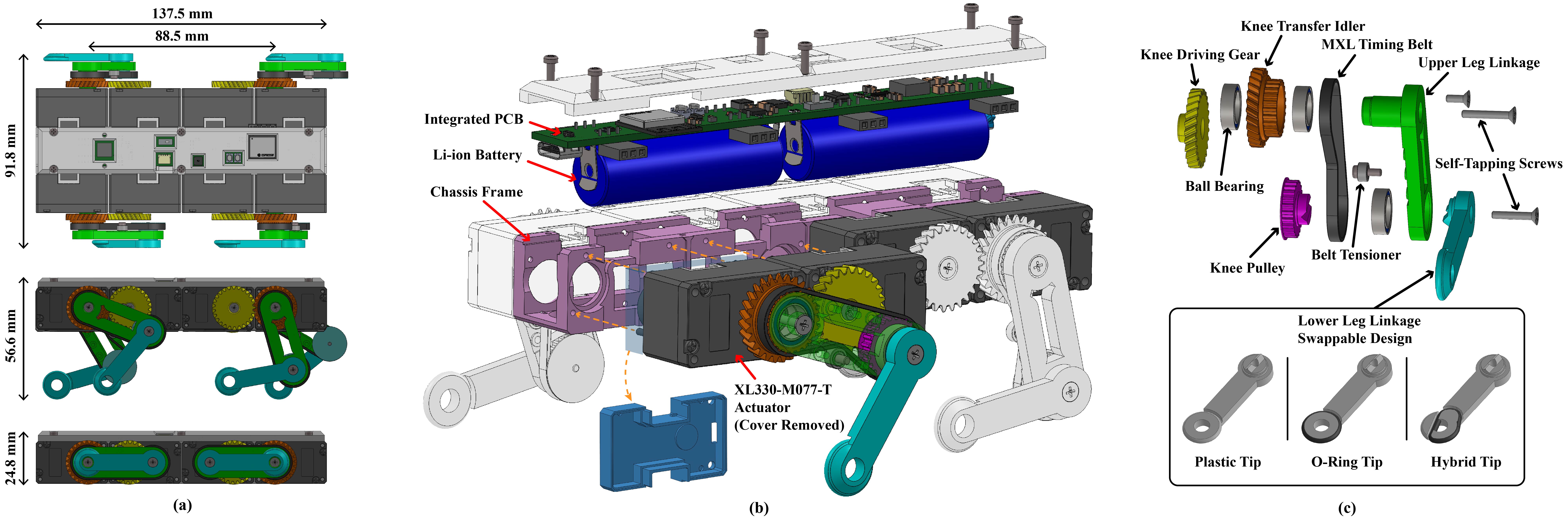}  
    \captionsetup{font=footnotesize}
    \captionsetup{belowskip=0pt}
    \caption{Mechanical design of MiNI-Q. (a) Overall dimensions in nominal and folded configurations. (b) Exploded view of the robot assembly. (c) Detail of MiNI-Q’s 2R leg mechanism enabling unbounded joint rotation. The lower leg linkage is designed to be easily swappable using a single screw, allowing for testing with different contact geometry and materials.}
    \label{fig:design}
\end{figure*}
\section{Methodology}

\subsection{Hardware Design}
MiNI-Q is a shorthand for "Miniature, No-wire, Intelligent Quadruped". As shown in Fig.~\ref{fig:design}, the robot features a re-designed chassis that adopts a similar, wire-free topology as its predecessor, Q8bot. An integrated printed circuit board (PCB) is mounted on top of the chassis frame, and 8 DYNAMIXEL XL330-M077-T actuators are mounted directly to the frame after removing their back covers and output horns. 4 identical leg assemblies are then mounted to their respective actuator pairs to form the complete robot. Table~\ref{tab:specs} compares the physical specifications of Q8bot and MiNI-Q. With all actuators mounted horizontally, the overall folded height is smaller than that of Q8bot. 

As part of the Q8bot design evolution, MiNI-Q reuses the same DYNAMIXEL actuators. With both electrical and mechanical design files released as open-source, users who have previously built Q8bot can upgrade to MiNI-Q for a fraction of the original system cost. The 2R leg mechanism of MiNI-Q uses accessible, off-the-shelf hardware combined with components fabricated on a hobby-grade FDM printer using PLA filament. A 0.2mm stainless steel nozzle was used to increase in-plane resolution, resulting in improved geometric accuracy for features such as gears, pulleys, and servo horn teeth.

\begin{table}[t]
\centering
\captionsetup{justification=centering}
\caption{Physical specifications of Q8bot and MiNI-Q.}
\label{tab:specs}
\begin{tabular}{|l|>{\centering\arraybackslash}m{1.5cm}>{\centering\arraybackslash}m{2.5cm}|}
\hline
\rule{0pt}{10pt}Specifications & \textbf{Q8bot} & \textbf{MiNI-Q} \\ [4pt]
\hline
\rule{0pt}{10pt}Mass (g) & 220 & 240 \\ [4pt]
\rule{0pt}{5pt}Nominal Dimension (cm) & 14 $\times$ 7 $\times$ 7 & 13.8 $\times$ 9.2 $\times$ 5.7 \\ [4pt]
\rule{0pt}{5pt}Folded Dimension (cm) & 12 $\times$ 7 $\times$ 5 & 13.8 $\times$ 9.2 $\times$ 2.5 \\ [4pt]
\rule{0pt}{5pt}Body Length (cm) & 8 & 8.8 \\ [4pt]
\rule{0pt}{5pt}Maximum Clearance (cm) & 5.8 & 5.8 \\ [4pt]
\hline
\end{tabular}
\end{table}

\begin{figure}[t]
    \centering
    \includegraphics[width=0.48\textwidth]{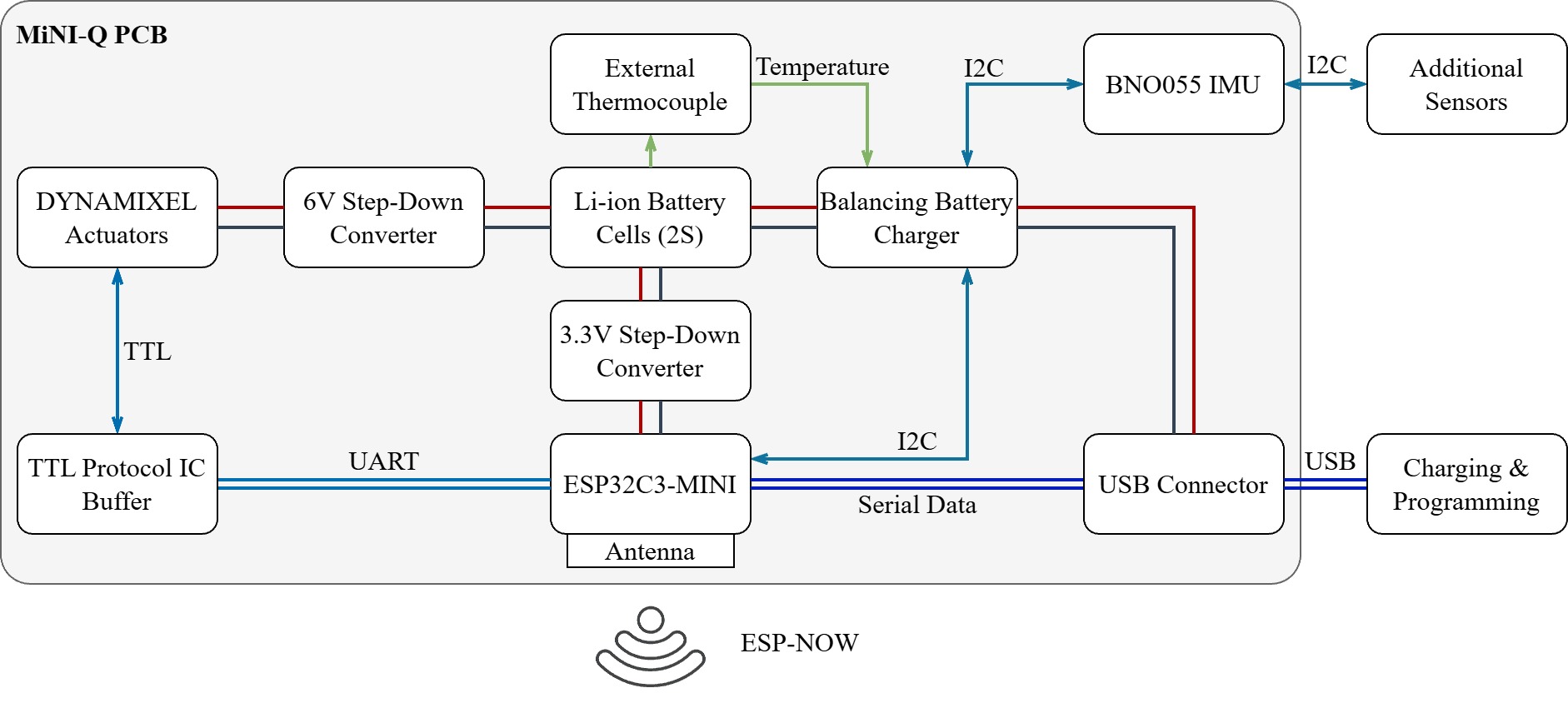}  
    \captionsetup{belowskip=0pt}
    \captionsetup{font=footnotesize}
    \caption{Electronics system overview of MiNI-Q}
    \label{fig:electronics}
\end{figure}

As shown in Fig.~\ref{fig:electronics}, MiNI-Q's integrated PCB represents a noticeable improvement over that of Q8bot. The off-the-shelf Seeed Studio XIAO ESP32C3 module was replaced by an ESP32C3-MINI system-in-package with a built-in PCB antenna, which further simplifies assembly. The battery configuration was updated from a 1S parallel arrangement to a 2S serial configuration. This allowed the use of a 6V step-down converter instead of a 5V boost converter to operate the actuators, increasing the rated stall torque by approximately 6\% and the no-load speed by 19\% \cite{xl330}. In addition, an onboard lithium-ion battery charger and a BNO055 inertial measurement unit (IMU) were added to make the robot fully self-contained. An I\textsuperscript{2}C bus connects the charger and the IMU to the microcontroller for data exchange, and a Qwiic connector was added to enable future expansion with additional sensors on the same bus \cite{BellQwiic}.

Due to its compact form factor and narrow, double-sided layout, MiNI-Q's custom integrated PCB is not well suited for manual pick-and-place assembly without professional reflow equipment. To improve accessibility and ensure design reproducibility, the components in the bill of materials were selected to be fully compatible with JLCPCB’s automated assembly service, eliminating the need for manual assembly or traditional turnkey quoting.

The most notable difference between Q8bot and MiNI-Q is the introduction of a joint-limit-free leg mechanism, shown in Fig.~\ref{fig:design}(c). The hip joint is directly driven by the first actuator, while the knee joint is remotely actuated through a transmission by a second actuator mounted rigidly relative to the first. This actuator drives a helical gear, which in turn moves a compound idler consisting of a matching helical gear stage and an integrated timing belt pulley. An MXL timing belt tensioned by a miniature ball bearing routes along the upper leg linkage, driving the lower leg linkage via the knee pulley. The resulting mechanism avoids physical overlap and interference between transmission stages, allowing both joints to rotate continuously. Unlike conventional coaxial five-bar leg designs, in which continuous joint rotation can only be achieved under coupled motion, our proposed design enables independent, unbounded rotation of each joint.

\subsection{Kinematic and Workspace Analysis}
\begin{figure}[t]
    \centering
    \includegraphics[width=0.48\textwidth]{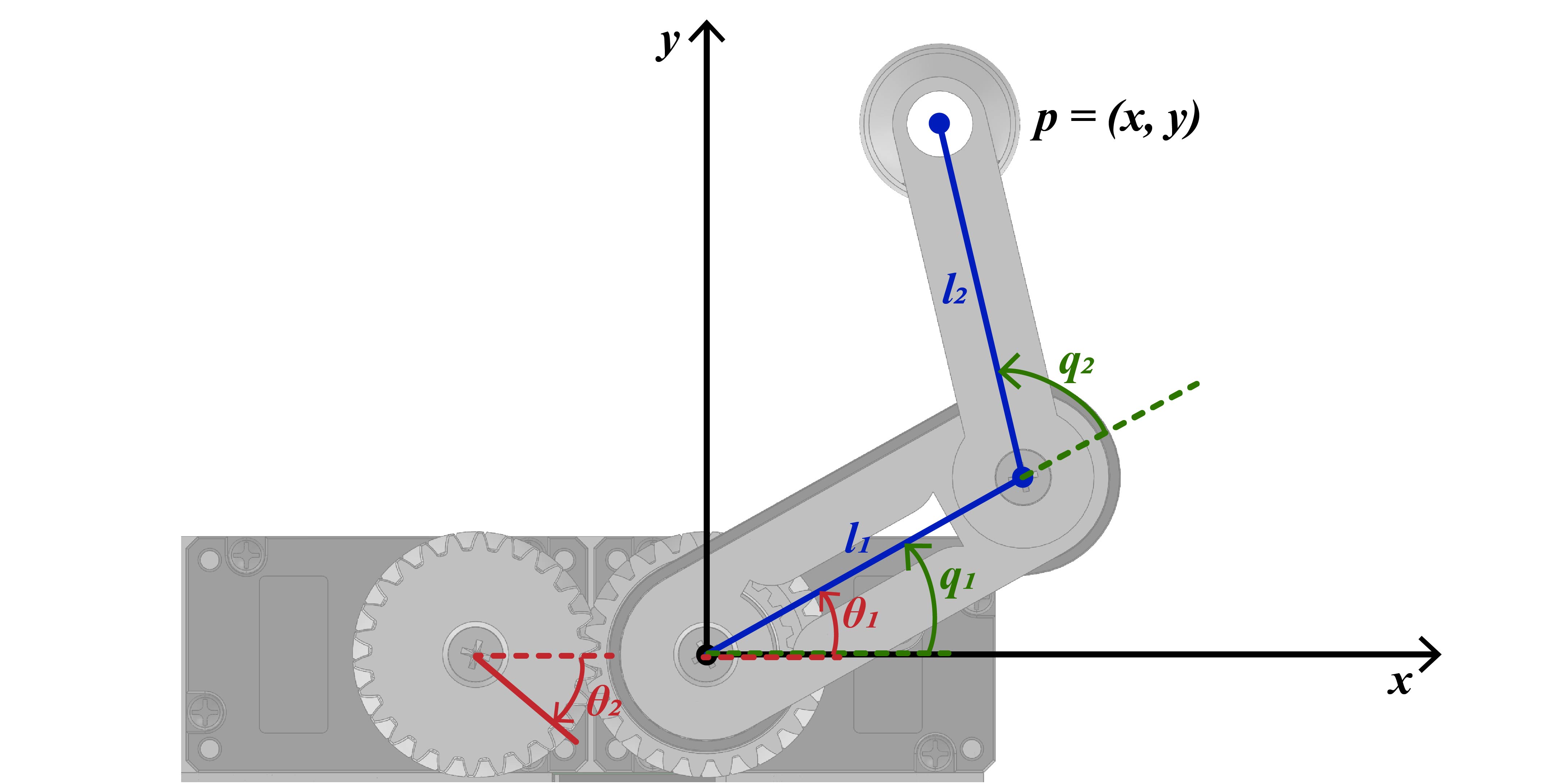}  
    \captionsetup{belowskip=0pt}
    \captionsetup{font=footnotesize}
    \caption{Kinematic diagram of MiNI-Q leg: A 2R linkage with custom joint-to-actuator mapping. \(q_1, q_2\) denote the kinematic joint angles, whereas \(\theta_1, \theta_2\) represent the corresponding actuator angles.}
    \label{fig:kinematics}
\end{figure}
The inverse kinematics of the proposed leg mechanism shown in Fig.\ref{fig:kinematics} can be computed by solving for the classical 2R manipulator solution in a theoretical joint space, then applying a joint-to-actuator mapping matrix. Given the leg lengths \(\boldsymbol{l_1, l_2}\) and desired position \(\boldsymbol{x, y}\) within the workspace, kinematic joint angles \(\boldsymbol{q_1, q_2}\) can be solved from the following equations

\begin{equation}
    \begin{bmatrix} 
        q_1 \\ 
        q_2 
    \end{bmatrix} = 
    \begin{bmatrix} 
        \operatorname{atan2}(y, x) - \operatorname{atan2}(l_2 \sqrt{1-D^2}, l_1 + l_2 D) \\ 
        \operatorname{atan2}(\pm \sqrt{1-D^2}, D) 
    \end{bmatrix}
\end{equation}

where
\begin{equation}
   D = \frac{x^2 + y^2 - l_1^2 - l_2^2}{2 l_1 l_2}
\end{equation}
The relationship between the kinematic joint angles \ \(\boldsymbol{q} = [q_1\; q_2]^\top\) and the actuator control angles \(\boldsymbol{\theta} = [\theta_1\; \theta_2]^\top\) is given by
\begin{equation}
    \boldsymbol{\theta} = A^{-1}\boldsymbol{q}, \quad
    A^{-1} =
    \begin{bmatrix}
    1 & 0 \\
    1 & 1
    \end{bmatrix}.
\end{equation}

\begin{figure}[t]
    \centering
    \includegraphics[width=0.48\textwidth]{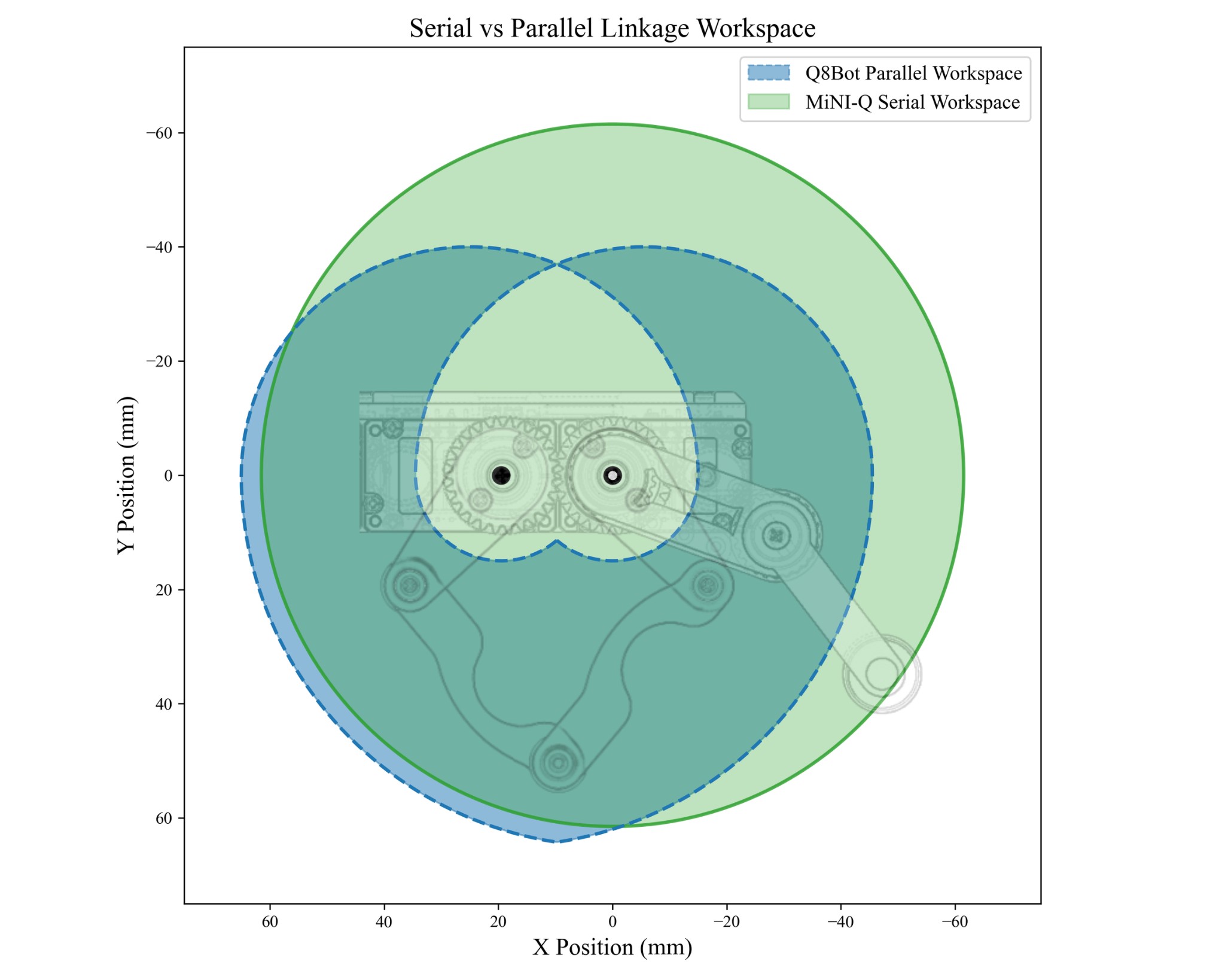}  
    \captionsetup{belowskip=0pt}
    \captionsetup{font=footnotesize}
    \caption{MiNI-Q's serial linkage workspace overlaid on top of Q8bot's parallel linkage workspace.}
    \label{fig:workspace}
\end{figure}

Without hard joint limits and equal link lengths, the reachable workspace of the leg mechanism becomes a circle with a radius equal to the total leg length at maximum extension. In comparison, Q8bot has a smaller leg workspace as seen in Fig. \ref{fig:workspace}. Additionally, due to the coupled kinematics of the parallel leg mechanism, Q8bot has a defined upright orientation with most of its reachable workspace below the chassis. However, the MiNI-Q's workspace allows for joint configurations above and below the chassis. This enables locomotion and movement that was previously not feasible with Q8bot, which is further explored in section \ref{sec:novel}.



\subsection{Jacobian Analysis}
The Jacobian of the proposed leg mechanism can be computed following a similar workflow. We first express the forward kinematics of the classical 2R manipulator as
\begin{equation}
\mathbf{p} =
\begin{bmatrix}
x \\
y
\end{bmatrix}
=
\begin{bmatrix}
l_1 \cos\theta_1 + l_2 \cos(\theta_1 + \theta_2) \\
l_1 \sin\theta_1 + l_2 \sin(\theta_1 + \theta_2)
\end{bmatrix}
\label{eq:fk_2r}
\end{equation}

The Jacobian in the theoretical joint space is obtained by direct differentiation of equation~\eqref{eq:fk_2r}:
\begin{equation}
\mathbf{J_{q}} =
\begin{bmatrix}
\displaystyle \frac{\partial x}{\partial q_1} &
\displaystyle \frac{\partial x}{\partial q_2} \\[6pt]
\displaystyle \frac{\partial y}{\partial q_1} &
\displaystyle \frac{\partial y}{\partial q_2}
\end{bmatrix}.
\label{eq:jacobian_q}
\end{equation}

Finally, the actuator-space Jacobian is computed by applying the joint-to-actuator mapping:
\begin{equation}
\mathbf{J}_{\theta}(\boldsymbol{\theta}) = \mathbf{J}_q(\mathbf{q})\,\mathbf{A}.
\label{eq:jacobian_theta}
\end{equation}

To quantify the dexterity of the leg mechanism, we utilize the established Yoshikawa manipulability index \cite{yoshikawa1985manipulability}, which is calculated as:
\begin{equation}
w(\boldsymbol{\theta}) = \sqrt{\det\!\big(\mathbf{J}(\boldsymbol{\theta})\,\mathbf{J}(\boldsymbol{\theta})^{\top}\big)}
\end{equation}

\begin{figure}[t]
    \centering
    \includegraphics[width=0.48\textwidth]{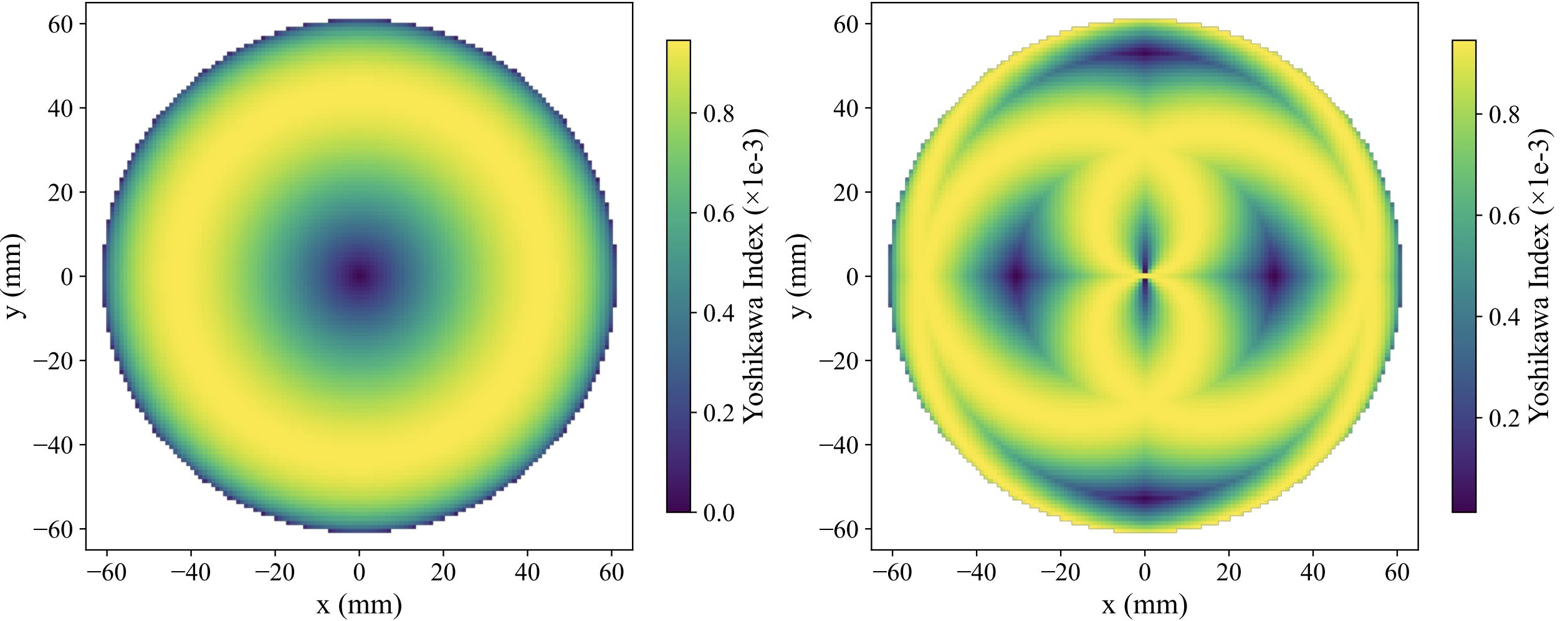}  
    \captionsetup{belowskip=0pt}
    \captionsetup{font=footnotesize}
    \caption{Yoshikawa manipulability across the workspace of MiNI-Q's 2R leg (left) and Q8bot's parallel leg mechanism (right). Lighter color represents higher manipulability.}
    \label{fig:manipulability}
\end{figure}

We plotted the Yoshikawa manipulability of the leg mechanisms of MiNI-Q and Q8bot for comparison in Fig.~\ref{fig:manipulability}. Q8bot has narrow bands of high manipulability alongside regions with severe directional constraints. In comparison, MiNI-Q has a radially symmetric manipulability with dead bands occurring on the workspace boundary and near the leg origin. This uniform dexterity ensures that MiNI-Q can maintain high locomotive ability regardless of leg configuration or gait selection. 

\subsection{Software and Control}



The MiNI-Q control architecture is based on Q8bot with several modifications. By incorporating an IMU, we migrated to the FreeRTOS framework to enable multi-task scheduling, allowing the robot to simultaneously handle teleoperated control and real-time sensor data collection without introducing latency or execution blocking. Additionally, we integrated a toggle between the DYNAMIXEL motors' position control and continuous velocity control modes. This control strategy takes full advantage of the infinite joint ranges, transforming the limbs to function as eccentric wheels to traverse highly variable terrain.
\section{Results}
\subsection{Performance} 
\label{sec:performance}

\begin{figure}[t]
\begin{subfigure}{0.5\textwidth}
    \centering
    \includegraphics[width=1\linewidth, height=6cm, keepaspectratio]{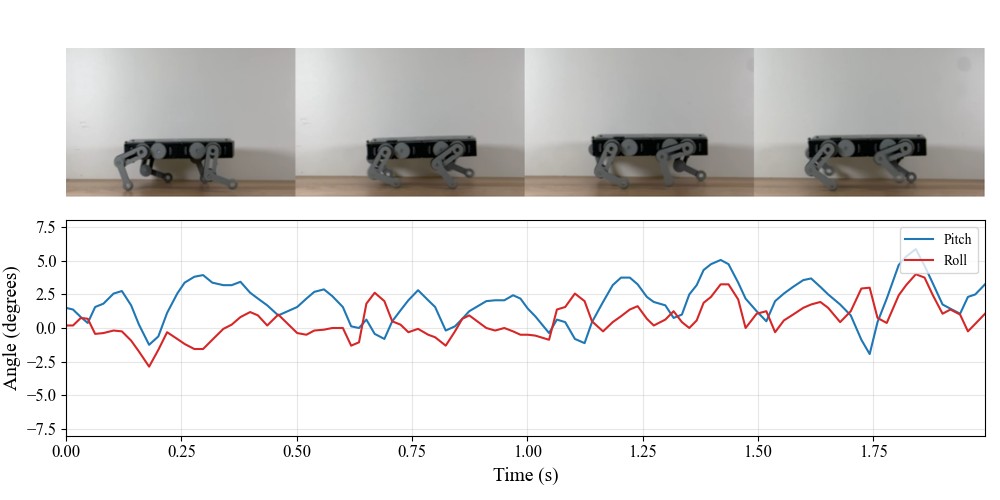} 
    \caption{}
    \label{fig:subim1}
\end{subfigure}
\begin{subfigure}{0.5\textwidth}
    \centering
    \includegraphics[width=1\linewidth, height=6cm, keepaspectratio]{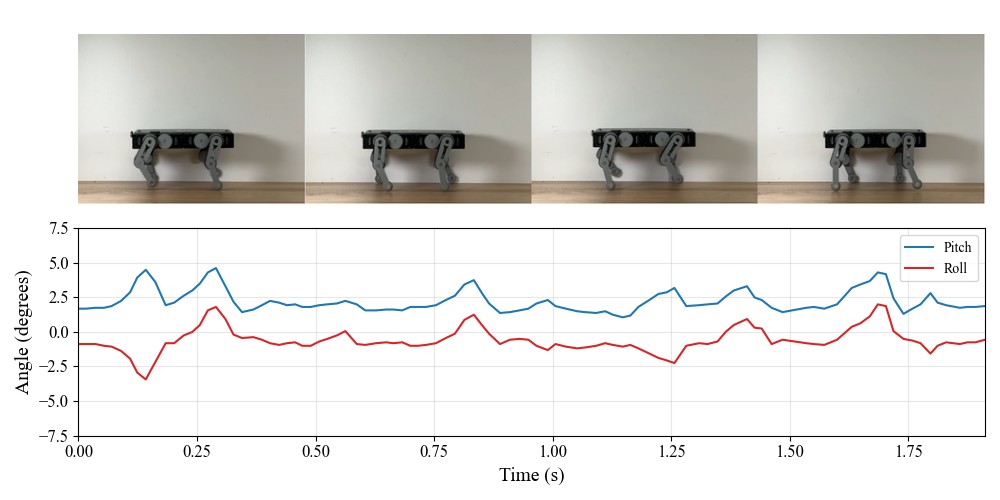}
    \caption{}
    \label{fig:subim2}
\end{subfigure}
\captionsetup{font=footnotesize}
\caption{Comparison of two trot gaits. 
(a) Fast trot. 
(b) High trot. 
Top: sequential snapshots of the robot over one gait cycle. 
Bottom: measured IMU pitch (blue) and roll (red) angles versus time.}
\label{fig:gaits-with-imu}
\end{figure}

To evaluate locomotion performance, we set up MiNI-Q to traverse in a straight line using different gait configurations. A video analysis software was used to extract the steady-state velocity of the robot \cite{tracker}, while IMU orientation angle and current draw were recorded at a frequency between 15-20 Hz. The IMU data for roll and pitch of MiNI-Q is shown for the fast and high trot gaits in Fig. \ref{fig:gaits-with-imu}.

Gait stability was quantified from the standard deviation of the roll and pitch angles measured from the IMU. Gait efficiency was measured using the Cost of Transport (COT) metric, derived from the robot mass $m$, steady-state velocity $v_{ss}$, average current consumption $i$, and input voltage $V$ (assuming a nominal voltage of 6.0 V):

\begin{equation}
    COT=\frac{Vi}{mg v_{ss}}
    \label{eq:cot}
\end{equation}

The results for various gaits are summarized in Table \ref{table:cot}.  All locomotion gaits exhibited stable performance, with body orientation deviating by a maximum of 2.19 degrees. The high-trot gait achieved the lowest COT of 7.1 at a speed of 0.16 m/s, while the fast trot reached a top speed of 0.46 m/s with a COT of 8.1. While MiNI-Q has a higher COT compared to other legged robots, the normalized speed is comparable to Q8bot, which outperforms most existing quadrupeds (Table \ref{table:comparison}).

\begin{table}[t]
\centering
\captionsetup{justification=centering}
\caption{Quadruped Performance Comparison}
\begin{tabular}{|l|>{\centering\arraybackslash}m{1.2cm} >{\centering\arraybackslash}m{1.5cm} >{\centering\arraybackslash}m{2cm}|}
\hline
\rule{0pt}{10pt}Robot &Top Speed &Normalized Top Speed &Cost of Transport (COT) \\ [4pt]
\hline
\rule{0pt}{10pt}\textbf{MiNI-Q}  &0.46 m/s &5.22 &7.1 \\ [4pt]
\rule{0pt}{10pt}\textbf{Q8Bot} \cite{Q8bot}  &0.43 m/s &5.38 &6.04\\ [4pt]
\rule{0pt}{10pt}Chen et al. \cite{ChenEtAl}  &0.52 m/s &4.4 &3.2\\ [4pt]
\rule{0pt}{10pt}Doggo \cite{Doggo}  &0.8 m/s &2.14 &3.2\\ [4pt]
\hline
\end{tabular}
\label{table:comparison}
\end{table}

\begin{table*}[t]
\centering
\captionsetup{justification=centering}
\caption{MiNI-Q COT Table}
\begin{tabular}{|l|>{\centering\arraybackslash}m{2cm} >{\centering\arraybackslash}m{2cm} >{\centering\arraybackslash}m{2cm} >{\centering\arraybackslash}m{2cm}|>{\centering\arraybackslash}m{2cm}>{\centering\arraybackslash}m{2cm}|}
\hline
\rule{0pt}{10pt}Gait Type &Steady State Velocity $V_{ss}$ &Normalized Velocity &Average Current Draw &Cost of Transport (COT) &Pitch Standard Deviation &Roll Standard Deviation \\ [4pt]
\hline
\rule{0pt}{10pt}Slow Trot  &0.12 m/s &1.36 &0.543 A &11.5 &2.15 &2.19\\ [4pt]
\rule{0pt}{10pt}Fast Trot  &\textbf{0.46 m/s} &5.22 &1.473 A &8.1 &1.57 &1.25\\ [4pt]
\rule{0pt}{10pt}High Trot  &0.16 m/s &1.81 &0.449 A &\textbf{7.1} &\textbf{0.73} &\textbf{0.80}\\ [4pt]
\rule{0pt}{10pt}Crawl  &0.03 m/s &0.32 &0.220 A &19.7 &0.85 &1.10\\ [4pt]
\hline
\end{tabular}
\label{table:cot}
\end{table*}

The difference in COT relative to Q8bot signifies an increased power requirement for locomotion, which can be attributed to the transition to the serial leg configuration. The parallel 5-bar linkage of Q8bot distributes the torque requirement per actuator, providing a mechanical advantage that is not present in MiNI-Q. Furthermore, the transmission drive for the distal serial link introduces energy loss due to internal friction and backlash. This marginal reduction in performance was an accepted design trade-off to achieve an unconstrained leg workspace, enabling gait generations that were not previously feasible with the parallel linkage legs.

\subsection{Novel Locomotion} 
\label{sec:novel}

\subsubsection{\textbf{Ledge Climbing}}
Several experiments were performed to demonstrate the locomotion capabilities previously restricted by the parallel linkage constraints of Q8bot. One such experiment is the ledge-climbing trial shown in Fig. \ref{fig:stairs}: the height of each step is 55 mm, which is 2.2 times the body height of MiNI-Q. The robot initiates the climb by extending each front leg onto the raised platform. Next, it executes a specialized gait to transition its center of mass over the ledge. Finally, the rear legs execute two full rotations to propel the entire chassis onto the platform.

\subsubsection{\textbf{Low Profile Crawl}}
Leveraging the unbounded workspace of MiNI-Q's legs, we implemented a low crawl gait by extending the legs outside of the chassis with a near-zero vertical offset as shown in Fig. \ref{fig:clearance}. This insect-like morphology allows MiNI-Q to navigate confined spaces as low as 45 mm, which would not have been achievable with the limited manipulability of Q8bot's legs or the finite joint ranges of most quadrupedal robots. 

\subsubsection{\textbf{Loose Substrate Navigation}}
To demonstrate MiNI-Q's adaptability, we conducted trials on loose gravel substrates. Using the fast trot gait, MiNI-Q's legs sank into the substrate due to the small foot contact area, resulting in an average speed of 0.038 m/s. However, by utilizing MiNI-Q's unbounded joint ranges and reconfiguring the DYNAMIXEL actuators to continuous rotation mode, we introduced a novel gait in which MiNI-Q begins with all legs spread outward horizontally, then continuously rotates the upper linkage joints as shown in Fig. \ref{fig:gravel}. By splaying the legs flat, MiNI-Q drastically increases its effective contact area, distributing its weight to prevent sinking into the substrate. Using this army-crawl configuration, MiNI-Q achieved an average speed of 0.13 m/s.

\subsection{Other Capabilities}

\begin{figure}
    \centering
    \includegraphics[width=1\linewidth]{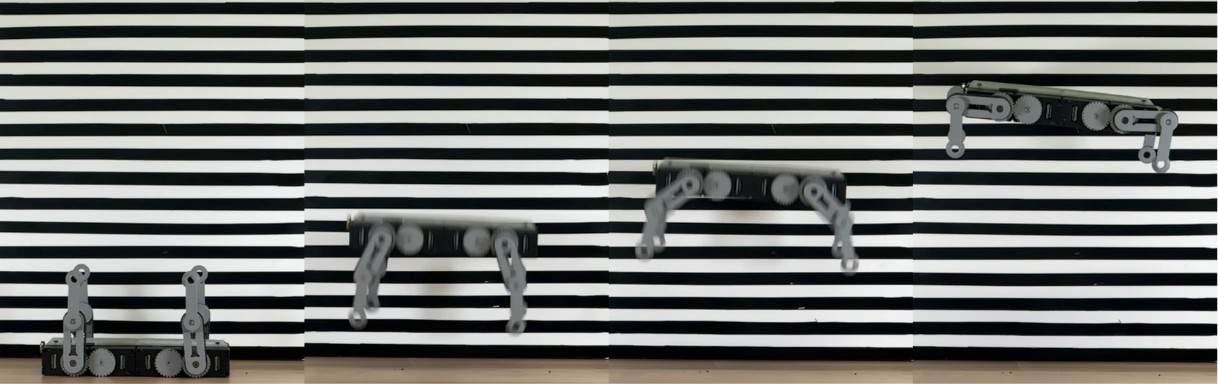}
    \captionsetup{font=footnotesize}
    \caption{MiNI-Q jumping where each black and white strip is 10 mm. Peak height is 220 mm.}
    \label{fig:jump}
\end{figure}

Beyond traversing complex terrain and obstacles, MiNI-Q achieves dynamic capabilities that remain out of reach for most traditional, workspace-constrained quadruped morphologies. While its predecessor, Q8bot, demonstrated a peak jump height of 70 mm \cite{Q8bot}, MiNI-Q can achieve a peak height of 220 mm as seen in Fig. \ref{fig:jump}. This represents approximately 9 times the robot's nominal chassis height. This dramatic performance leap is made possible by upgraded power electronics coupled with the unconstrained leg workspace. Rather than executing a standard contact-based jump, MiNI-Q initiates the jump sequence with its legs extended upward and the body resting on the ground; the actuators then rapidly drive the limbs downward, applying an impulsive high-velocity strike against the ground to maximize initial reaction forces and optimize vertical acceleration.

\begin{figure*}[t]
    \centering

    \begin{subfigure}[b]{1.0\linewidth}
        \centering
        \includegraphics[width=1\linewidth]{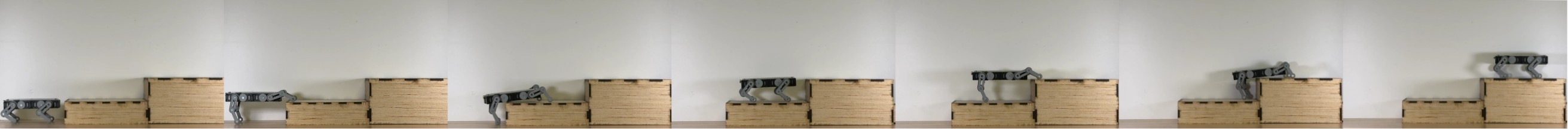}
        \caption{Stair Climbing}
        \label{fig:stairs}
    \end{subfigure}
    
    \par\bigskip 
    
    \begin{subfigure}[b]{1.0\linewidth}
        \centering
        \includegraphics[width=1\linewidth]{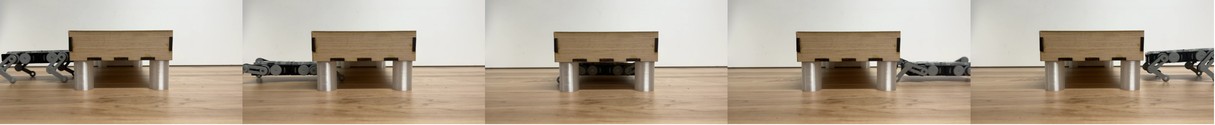}
        \caption{Low Clearance}
        \label{fig:clearance}
    \end{subfigure}
    
    \par\bigskip 

    \begin{subfigure}[b]{1.0\linewidth}
        \centering
        \includegraphics[width=1\linewidth]{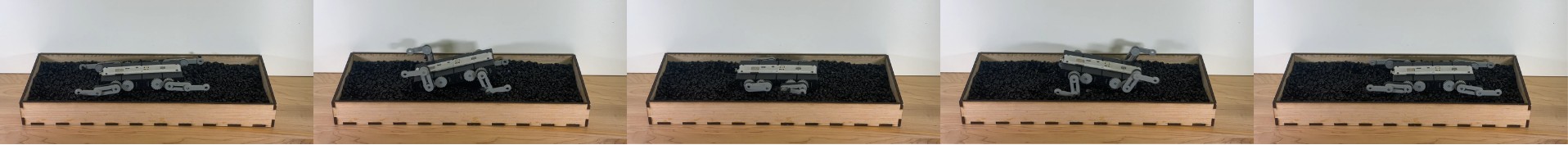}
        \caption{Gravel Traversal}
        \label{fig:gravel}
    \end{subfigure}
    
    \par\bigskip

    \begin{subfigure}[b]{1.0\linewidth}
        \centering
        \includegraphics[width=1\linewidth]{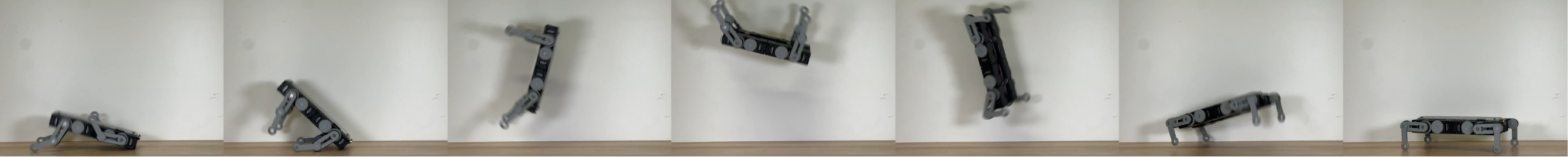}
        \caption{Backflip Maneuver}
        \label{fig:backflip}
    \end{subfigure}
    \captionsetup{font=footnotesize}
    \caption{Demonstrations of MiNI-Q locomotion capabilities. 
(a) Stair climbing up steps $2.2\times$ the robot’s body height. 
(b) Low-clearance traversal through a $45\,\mathrm{mm}$ gap. 
(c) Locomotion on pebbled terrain using the continuous rotation gait. 
(d) Jumping backflip maneuver.}
    \label{fig:vertical_gaits}
\end{figure*}

In addition to vertical jumping, MiNI-Q can also execute a dynamic backflip as seen in Fig. \ref{fig:backflip}. This maneuver was unattainable for Q8bot due to the restricted workspace of the 5-bar linkage legs. Similar to the jump sequence, by optimizing the leg trajectories and the final joint configurations at liftoff, the legs apply a coordinated wrench to the chassis, generating the necessary angular momentum for the robot to complete a full in-air rotation before landing.

\begin{figure}
    \centering
    \includegraphics[width=1\linewidth]{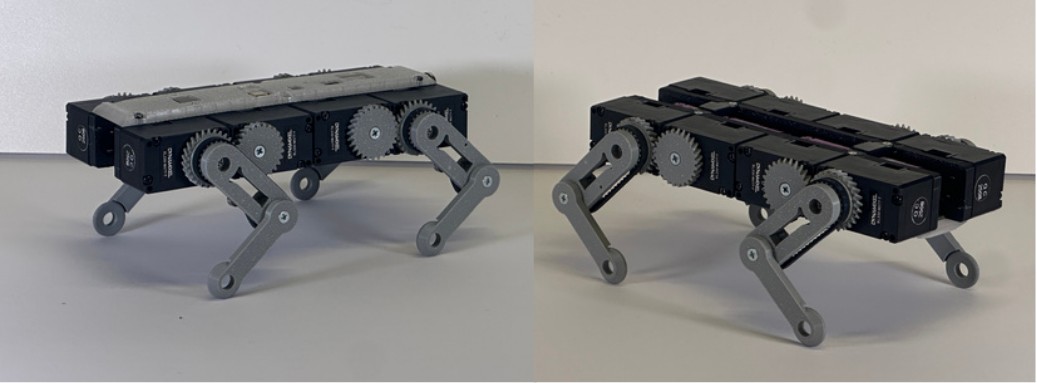}
    \captionsetup{font=footnotesize}
    \caption{MiNI-Q standing in upright position (Left) and inverted position (Right). Note the location of the white PCB cover.}
    \label{fig:inverted}
\end{figure}

Lastly, utilizing its onboard IMU sensor to detect orientation, MiNI-Q leverages its expanded workspace to achieve instantaneous flip recovery. Unlike traditional quadrupeds that require complex, multi-step self-righting sequences when inverted, MiNI-Q can simply pivot its limbs 180\degree\ within 0.5 seconds as shown in Fig. \ref{fig:inverted}. Because the leg mechanism is symmetric and unconstrained, the robot can immediately resume any locomotive gait from this inverted state, rendering the platform effectively orientation-agnostic.

\section{Discussion}
\subsection{Design Choices} 

Unlike quadrupeds with bulky top-mounted components or asymmetric knee offsets, MiNI-Q is compact and symmetric about its horizontal plane. This geometric balancing ensures that the vertical clearance between the chassis and the ground remains consistent in both upright and inverted states, which enables all gaits to perform with equal efficacy regardless of the robot's orientation.

Furthermore, modularity was integrated into the leg architecture by making the distal link easily swappable with a single screw, as seen in Fig. \ref{fig:design}(c). This facilitates rapid adjustments to the foot's geometric profile across various traversal experiments. To complement this, a hybrid foot tip was developed, featuring distinct rubber and plastic contact surfaces; by transitioning between an ``elbow-in" or ``elbow-out" posture, the robot can actively select which material interfaces with the ground, allowing for high performance locomotion depending on the terrain environment.

\subsection{Limitations} 
While the experimental results validate MiNI-Q's mechanical versatility, several constraints remain. First, locomotion trials were limited to open-loop control, with joint trajectories pre-calculated on a host laptop. This architecture was necessitated by the processing constraints of the onboard ESP32-C3 microcontroller, which can experience significant scheduling delays within the FreeRTOS framework when concurrently managing ESP-NOW communication with the host, half-duplex DYNAMIXEL servo actuation, and high-frequency sensor data polling. Although these latency issues were mitigated through code optimization and offboard computation, this approach limits the robot's ability to respond to abrupt disturbances or unmodeled terrain variations.

Second, the fidelity of the internal current sensors of the DYNAMIXEL actuators remains a point of investigation. These sensors may be subject to noise that affects the precision of the COT calculation. Furthermore, while the bus voltage was assumed to be constant at 6.0 V for simplification, fluctuations in battery voltage and back-EMF generated by the actuators can introduce minor discrepancies between the calculated and actual power draw.


\subsection{Future Work}
 
To address the current limitations in control and data accuracy, future work will involve a comprehensive overhaul of the integrated PCB design. The primary upgrade involves transitioning to an ESP32-S3 dual-core microcontroller to facilitate the strategic partitioning of time-critical tasks to eliminate scheduling conflicts. Additionally, to ensure high-accuracy metrics, we plan to integrate current-shunt monitors (e.g. the INA260) into the power bus. Finally, peripherals such as UART and SPI will be exposed for integration with external modules such as a Single-Board Computer (SBC), enabling the transition to advanced control architectures using middleware systems such as the Robot Operating System (ROS 2).

Leveraging this enhanced computational and sensing hardware, future experiments will focus on the implementation of closed-loop locomotion strategies. We intend to investigate the feasibility of proprioceptive contact sensing while maintaining the simplicity of a no-wire design. For instance, by using high-fidelity current data from the INA260 sensors, we can analyze transient torque impulses and deviations in motor velocity to infer contact events.

Looking ahead, the maturity of the MiNI-Q hardware provides a versatile foundation to pursue distinct, multi-disciplinary research directions. One research avenue focuses on leveraging Reinforcement Learning (RL) frameworks to train robust locomotion policies. Prior work by Fu et al. demonstrated that learning to minimize mechanical energy promotes the emergence of natural, efficient locomotion gaits across diverse terrain settings \cite{RLEnergy}. Building upon this, we intend to implement a learning and sim-to-real framework for MiNI-Q to realize non-intuitive, energy-optimal locomotion patterns that a manually tuned controller might overlook.

A parallel research direction will pivot towards utilizing MiNI-Q as a highly scalable platform for swarm and collaborative robotics. Given the robot's low-cost, compact footprint, and inherent physical resilience, it is uniquely suited for deployment in large, decentralized fleets navigating through terrain-variant environments. Future efforts will leverage the onboard SBC alongside Ultra-Wideband (UWB) modules and RGB cameras to implement precise localization and ad-hoc communication networks, allowing a fleet of MiNI-Q quadrupeds to coordinate without a centralized controller. We intend to explore collective behaviors such as distributed mapping of unstructured environments, cooperative object interaction, and fleet formation. Because an individual MiNI-Q can instantaneously recover from flips or falls, the entire swarm gains a high level of operational robustness, ensuring that localized terrain failures do not compromise the collective task objective. Furthermore, the unique kinematic capabilities of MiNI-Q can be exploited to realize novel tasks within the area of swarm and collaborative robotics.
\section{Conclusion} 

MiNI-Q extends the wire-free architecture set forth by Q8bot. Its unique, independently actuated serial leg mechanism creates an unlimited workspace, opening up the possibility of novel gaits and locomotion. Such examples are stair-climbing, backflipping, continuous rotation, and low clearance crawling. Experimental results validated the robot's performance, demonstrating stable locomotion with a normalized top speed of 5.22 body lengths per second and a COT comparable to existing miniature platforms. While the transition to a serial linkage introduced minor trade-offs in energy efficiency due to transmission losses, this was outweighed by the functional benefits of an unconstrained workspace. The wire-free architecture and open-source design lower the barrier to entry for research in legged robotics, providing a robust and accessible platform for studying novel gait generation and control strategies.

\section*{Acknowledgment}
The authors would like to thank Jessica Yang for her help in creating the spatial calibration grid, which was used in various experiments. Additionally, they would like to thank the UCLA Engineering Makerspace for providing resources and equipment to build parts of the experimental setup.

\bibliographystyle{ieeetr}
\bibliography{main}

@article{QuadrupedReview,
  title={A Novel Review on Quadruped Robots Design Variants, Gait Modulation, and Motion Planning Schemes},
  author={Azeez, SA and Mandava, Ravi Kumar and Naik, Nenavath Srinivas},
  journal={Journal of Field Robotics},
  year={2025},
  publisher={Wiley Online Library}}

@INPROCEEDINGS{LegDisney,
  author={Gim, Kevin G. and Kim, Joohyung and Yamane, Katsu},
  booktitle={2018 IEEE International Conference on Robotics and Automation (ICRA)}, 
  title={Design of a Serial-Parallel Hybrid Leg for a Humanoid Robot}, 
  year={2018},
  volume={},
  number={},
  pages={6076-6081},
  doi={10.1109/ICRA.2018.8460733}}

@article{LegOncilla,
  title={Oncilla robot: a versatile open-source quadruped research robot with compliant pantograph legs},
  author={Spr{\"o}witz, Alexander T and Tuleu, Alexandre and Ajallooeian, Mostafa and Vespignani, Massimo and M{\"o}ckel, Rico and Eckert, Peter and D'Haene, Michiel and Degrave, Jonas and Nordmann, Arne and Schrauwen, Benjamin and others},
  journal={Frontiers in Robotics and AI},
  volume={5},
  pages={67},
  year={2018},
  publisher={Frontiers Media SA}}

@INPROCEEDINGS{ActuatorBEAR,
  author={Zhu, Taoyuanmin and Hooks, Joshua and Hong, Dennis},
  booktitle={2019 IEEE/ASME International Conference on Advanced Intelligent Mechatronics (AIM)}, 
  title={Design, Modeling, and Analysis of a Liquid Cooled Proprioceptive Actuator for Legged Robots}, 
  year={2019},
  volume={},
  number={},
  pages={36-43},
  doi={10.1109/AIM.2019.8868596}}

@article{RLReview,
  title={Learning-based legged locomotion: State of the art and future perspectives},
  author={Ha, Sehoon and Lee, Joonho and van de Panne, Michiel and Xie, Zhaoming and Yu, Wenhao and Khadiv, Majid},
  journal={The International Journal of Robotics Research},
  volume={44},
  number={8},
  pages={1396--1427},
  year={2025},
  publisher={Sage Publications Sage UK: London, England}
}

@article{MechanismReview,
  title={Mechanism, actuation, perception, and control of highly dynamic multilegged robots: A review},
  author={He, Jun and Gao, Feng},
  journal={Chinese Journal of Mechanical Engineering},
  volume={33},
  number={1},
  pages={79},
  year={2020},
  publisher={Springer}}

@INPROCEEDINGS{MrongaParallel,
  author={Mronga, Dennis and Kumar, Shivesh and Kirchner, Frank},
  booktitle={2022 International Conference on Robotics and Automation (ICRA)}, 
  title={Whole-Body Control of Series-Parallel Hybrid Robots}, 
  year={2022},
  volume={},
  number={},
  pages={228-234},
  doi={10.1109/ICRA46639.2022.9811616}}

@ARTICLE{Lee_AiDIN,
  author={Lee, Yoon Haeng and Lee, Young Hun and Lee, Hyunyong and Kang, Hansol and Lee, Jun Hyuk and Phan, Luong Tin and Jin, Sungmoon and Kim, Yong Bum and Seok, Dong-Yeop and Lee, Seung Yeon and Moon, Hyungpil and Koo, Ja Choon and Choi, Hyouk Ryeol},
  journal={IEEE Transactions on Industrial Electronics}, 
  title={Development of a Quadruped Robot System With Torque-Controllable Modular Actuator Unit}, 
  year={2021},
  volume={68},
  number={8},
  pages={7263-7273},
  doi={10.1109/TIE.2020.3007084}}

@ARTICLE{Grimminger_Solo,
  author={Grimminger, Felix and Meduri, Avadesh and Khadiv, Majid and Viereck, Julian and Wüthrich, Manuel and Naveau, Maximilien and Berenz, Vincent and Heim, Steve and Widmaier, Felix and Flayols, Thomas and Fiene, Jonathan and Badri-Spröwitz, Alexander and Righetti, Ludovic},
  journal={IEEE Robotics and Automation Letters}, 
  title={An Open Torque-Controlled Modular Robot Architecture for Legged Locomotion Research}, 
  year={2020},
  volume={5},
  number={2},
  pages={3650-3657},
  doi={10.1109/LRA.2020.2976639}}

@misc{BillingsBDPatent,
  author = {Devin Billings and Steven Potter},
  title = {Electrical transfer assemblies for robotic devices},
  howpublished = {U.S. Patent US20240181661A1},
  year = {2024},
  month = jun,
}

@article{xu2023design,
  title={Design and experiments of a human-leg-inspired omnidirectional robotic leg},
  author={Xu, Yuze and Luo, Zirong and Bai, Xiangjuan and Xie, Huixiang and Zhu, Yiming and Chen, Shanjun and Shang, Jianzhong},
  journal={Journal of Bionic Engineering},
  volume={20},
  number={6},
  pages={2570--2589},
  year={2023},
  publisher={Springer}}

@INPROCEEDINGS{Doggo,
  author={Kau, Nathan and Schultz, Aaron and Ferrante, Natalie and Slade, Patrick},
  booktitle={2019 International Conference on Robotics and Automation (ICRA)}, 
  title={Stanford Doggo: An Open-Source, Quasi-Direct-Drive Quadruped}, 
  year={2019},
  volume={},
  number={},
  pages={6309-6315},
  doi={10.1109/ICRA.2019.8794436}}

@INPROCEEDINGS{Lywal,
  author={Xue, Yongjiang and Yuan, Xichen and Wang, Yuhai and Yang, Yang and Lu, Siyu and Zhang, Bo and Lai, Juezhu and Wang, Jianming and Xiao, Xuan},
  booktitle={2021 IEEE International Conference on Robotics and Automation (ICRA)}, 
  title={Lywal: a Leg-Wheel Transformable Quadruped Robot with Picking up and Transport Functions}, 
  year={2021},
  volume={},
  number={},
  pages={2935-2941},
  doi={10.1109/ICRA48506.2021.9561128}}

@ARTICLE{RHex-T3,
  author={Sun, Chunhu and Yang, Guiyu and Yao, Senge and Liu, Qi and Wang, Jianming and Xiao, Xuan},
  journal={IEEE/ASME Transactions on Mechatronics}, 
  title={RHex-T3: A Transformable Hexapod Robot With Ladder Climbing Function}, 
  year={2023},
  volume={28},
  number={4},
  pages={1939-1947},
  doi={10.1109/TMECH.2023.3276756}}

@INPROCEEDINGS{NaBi,
  author={Ghassemi, Sepehr and Hong, Dennis},
  booktitle={2018 15th International Conference on Ubiquitous Robots (UR)}, 
  title={Investigation of a Novel Continuously Rotating Knee Mechanism for Legged Robots}, 
  year={2018},
  volume={},
  number={},
  pages={130-135},
  doi={10.1109/URAI.2018.8441802}}

@INPROCEEDINGS{Q8bot,
  author={Wu, Yufeng and Hong, Dennis},
  booktitle={2025 IEEE/RSJ International Conference on Intelligent Robots and Systems (IROS)}, 
  title={Design of Q8bot: A Miniature, Low-Cost, Dynamic Quadruped Built with Zero Wires}, 
  year={2025},
  volume={},
  number={},
  pages={8820-8825},
  doi={10.1109/IROS60139.2025.11246322}}

@manual{xl330,
  title        = {XL330-M077-T e-Manual},
  organization = {ROBOTIS},
  year         = {2026},
  note         = {Online technical manual},
  url          = {https://emanual.robotis.com/docs/en/dxl/x/xl330-m077/}
}

@incollection{BellQwiic,
  title={Introducing Qwiic and STEMMA QT},
  author={Bell, Charles},
  booktitle={Beginning IoT Projects: Breadboard-less Electronic Projects},
  pages={217--258},
  year={2021},
  publisher={Springer}
}

@article{yoshikawa1985manipulability,
  title={Manipulability of robotic mechanisms},
  author={Yoshikawa, Tsuneo},
  journal={The international journal of Robotics Research},
  volume={4},
  number={2},
  pages={3--9},
  year={1985},
  publisher={Sage Publications Sage CA: Thousand Oaks, CA}}

@article{tracker,
  title={Innovative uses of video analysis},
  author={Brown, Douglas and Cox, Anne J},
  journal={The Physics Teacher},
  volume={47},
  number={3},
  pages={145--150},
  year={2009},
  publisher={American Association of Physics Teachers}
}

@misc{RLEnergy,
      title={Minimizing Energy Consumption Leads to the Emergence of Gaits in Legged Robots}, 
      author={Zipeng Fu and Ashish Kumar and Jitendra Malik and Deepak Pathak},
      year={2021},
      eprint={2111.01674},
      archivePrefix={arXiv},
      primaryClass={cs.RO},
      url={https://arxiv.org/abs/2111.01674}, 
}

@INPROCEEDINGS{ChenEtAl,
  author={Chen, Fuchen and Tao, Weijia and Aukes, Daniel M.},
  booktitle={2023 IEEE/RSJ International Conference on Intelligent Robots and Systems (IROS)}, 
  title={Development of A Dynamic Quadruped with Tunable, Compliant Legs}, 
  year={2023},
  volume={},
  number={},
  pages={495-502},
  doi={10.1109/IROS55552.2023.10342283}}
\end{document}